\documentclass[twocolumn,10pt]{asme2e}
\usepackage[pass,paperwidth=8.5in,paperheight=11in]{geometry}
\usepackage{amsmath}
\usepackage{graphicx}
\usepackage{titlesec}
\usepackage{indentfirst}
\confshortname{IDETC/CIE 2017}
\conffullname{the ASME 2017 International Design Engineering Technical Conferences \&\\
              Computers and Information in Engineering Conference}

\confdate{August 6-9}
\confyear{2017}
\confcity{Cleveland}
\confcountry{USA}

\papernum{DETC2017-68082}

\title{A LOBSTER-INSPIRED HYBRID ACTUATOR\\WITH RIGID AND SOFT COMPONENTS}
\author{Yaohui Chen, Sing Le, Qiao Chu Tan, Oscar Lau, Chaoyang Song\thanks{Address all correspondence to this author at chaoyang.song@monash.edu} 
    \affiliation{Sustainable and Intelligent Robotics Group\\
    Department of Mechanical Engineering\\
	Monash University\\
	Clayton, Victoria, 3800\\
	Australia\\
    }
}

\begin{document}

\maketitle    
\titlespacing*{\section} {0pt}{9pt}{0pt}  
\titlespacing*{\subsection} {0pt}{9pt}{0pt}
\titlespacing*{\subsubsection} {0pt}{9pt}{0pt}
\begin{abstract}
\textit{Soft actuators} have drawn significant attentions from researchers with an inherently compliant design to address the safety issues in physical human-robot interactions. However, they are also vulnerable and pose new challenges in the design, fabrication, and analysis due to their inherent material softness. In this paper, a novel \textit{hybrid actuator} design is presented with bio-inspirations from the lobster, or crustaceans in a broader perspective. We enclose a soft chamber with rectangular cross-section using a series of articulated rigid shells to produce bending under pneumatic input. By mimicking the shell pattern of lobsters' abdomen, foldable rigid shells are designed to provide the soft actuator with full protection throughout the motion range. The articulation of the rigid shells predefines the actuator's bending motions. As a result, the proposed design enables one to analyze this hybrid actuator with simplified quasi-static models and rigid-body kinematics, which are further validated by mechanical tests. This paper demonstrates that the proposed hybrid actuator design is capable of bridging the major design drawbacks of the entirely rigid and soft robots while preserving their engineering merits in performance.
\begin{flushleft}
\end{flushleft}
\end{abstract}

\section{INTRODUCTION}
Most classical robotic actuators are rigidly built with advantages in efficiency, robustness, and precision. Recent development in soft actuators has emerged as a dynamically evolving research field by utilizing the material softness for robotic actuation, leading to robots that are life-like, compliant, and light-weight with reduced fabrication cost \cite{Kim2013SoftRobotics, Laschi2016SoftAbilities, Trimmer2015SoftSize}. In soft pneumatic actuator (SPA) designs like PneuNets, a series of air chambers and corridors inside produces controllable motion through material deformation under pneumatic actuation \cite{Ilievski2011SoftChemists, Shepherd2011MultigaitRobot, Sun2013CharacterizationActuators}. One can generate programmable actuation by modifying the design of the air chambers and material softness for linear, bending and twisting motions. Applications of SPAs can be found in human-assistive robotics devices \cite{Polygerinos2015SoftRehabilitation, Polygerinos2013TowardsRehabilitation}, adaptive locomotion \cite{Tolley2014ARobot} and delicate object manipulation \cite{Paek2015MicroroboticMicrotubes, Ilievski2011SoftChemists}. 

However, soft actuators suffer from some inherent disadvantages including \textit{vulnerability to ruptures}, \textit{complexity in fabrication}, and \textit{difficulty in analysis}. SPAs experience a reduction in wall thickness upon actuation, resulting a higher chance of perforation on these inflated thin and soft walls during physical contact with rough particles or surfaces. Solutions to circumvent this has been explored from constraining the radial expansion with fibers\cite{Roche2014AMaterial, Connolly2015MechanicalAngle}, to using a more robust outer layer recently such as an outer shell with openings \cite{Agarwal2016StretchableDevices, Memarian2015ModellingMuscles} and origami \cite{Paez2016DesignReinforcement}. Moreover, as SPAs' mechanical performance depends a lot on the geometric parameters, small differences in fabrication sequence, timing and environment can significantly alter the individual results \cite{Agarwal2016StretchableDevices}, making it difficult to maintain an acceptable repeatability in production with a complex fabrication process. Furthermore, the continuous deformation of soft actuators can no longer be described by rigid-body kinematics, and the complex non-linear stress-strain relationships of common soft materials require a more complex hyperelastic material law. While soft actuators with specific applications have been investigated using numerical tools \cite{Moseley2016ModelingMethod} or quasi-static analytic models \cite{Polygerinos2015ModelingActuators,  Wang2017InteractionActuators}, an explicit analysis of their mechanical performance is still challenging.

\begin{figure}[htbp]
    \begin{center}
    \includegraphics[width=1\linewidth]{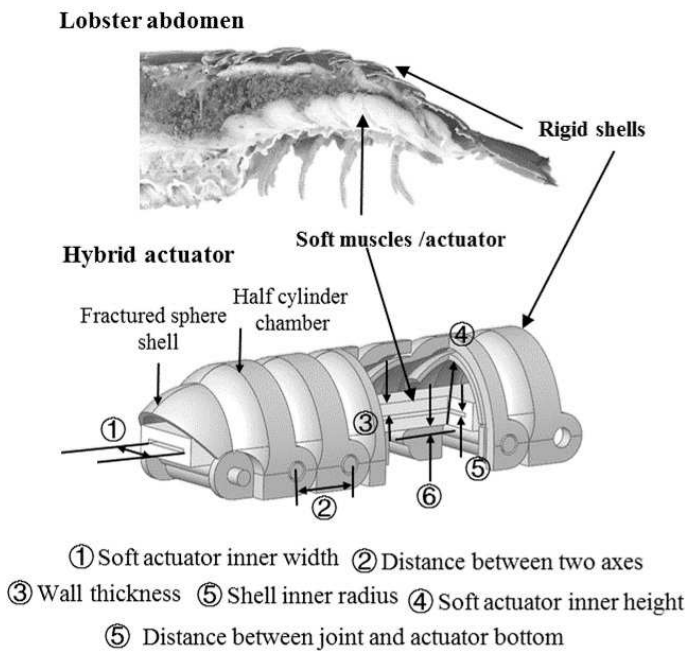}
    \end{center}
    \caption{Bio-inspiration from anatomical structure of lobster's abdomen and geometric parameters that can influence the behavior.}
    \label{fig:Bioinspire}
\end{figure}

To overcome the drawbacks mentioned above with enhanced robustness in mechanical performance, a new type of bending actuators, comprising both soft and rigid components, is proposed in this work. Crustaceans such as lobsters swim under water through the flapping motion from their abdomen with a backward propulsion as a startle-escape response \cite{Newland1988Swimming, Arnott1998Tail-flipCrangon}. These tail-flips are brought about by soft muscle's rapid alternate extensions and flexions within their abdomen, pushing or dragging rigid exoskeletons to move in a rotary trajectory \cite{Newland1992EscapeLobster}. The anatomical structure of lobster's abdomen leads to the design of the proposed hybrid actuator including rigid shell segments outside and a soft actuator inside, as shown in Fig. \ref{fig:Bioinspire}. The soft chamber inside provides necessary actuation like the abdomen muscle of a lobster, while the rigid shell segments serve multiple purposes to combat the disadvantages of soft actuators with preserved advantages in rigid robotics \cite{Chen2017ARehabilitation, Chen2017AComponents}.

In the rest of this paper, section 2 explains the design principles and fabrication procedures of this hybrid actuator. In section 3, a quasi-static model is proposed to analyze the bending actuation and tip force exertion of this hybrid actuator, along with its forward kinematics. Experimental validation is carried out in section 4 to characterize the performance of this hybrid actuator. Conclusion and limitations of this paper are enclosed in the last part. 

\section{ACTUATOR DESIGN AND FABRICATION}
Unlike rigid actuators with limited compliance or soft actuators with infinite degrees of freedom, the serially hinged rigid shells create multiple degree-of-freedoms (DOFs) for the hybrid actuator to achieve a good compliance. This mechanism defines a precise bending trajectory for the hybrid actuator, making it comparable to a robotic manipulator with known forward kinematics during bending motion. As a result, this makes the hybrid actuator fundamentally different from most existing soft actuators whose continuous deformation is beyond the capacity of traditional robotic theory.  

We design a novel folding mechanism with inspirations from lobster's shell pattern to provide an improved protection to the soft chamber inside. As shown in Fig.\ref{fig:Bioinspire}, the fractured sphere shells are foldable into an adjacent segment's half cylinder chamber before actuation. Once actuated by pressurizing the soft chamber inside, the inflated soft actuator expands and pushes the edge of the fracture sphere shells to rotate around the joint axes to accomplish a bending motion. The half cylinder chambers along with the fractured sphere shells can fully envelop the soft chamber throughout the range of bending motion, and excessive local stretch is therefore completely avoided. The rigid shells also provide stronger mechanical constraints to the radial expansion of the soft chamber, enabling the hybrid actuator to sustain higher pressure with enhanced robustness. Moreover, as the soft chamber will conform to the internal geometry of the rigid shells, it provides the flexibility in soft chamber design, where a much simpler geometry can be used. For example, a rectangular soft chamber is adopted in this work. Other simple cross-sections, i.e. circle, triangle, can be utilized as well. The resultant soft chamber exhibits a simplified design and fabrication process without sacrificing the overall performance of the actuator, as will be shown later in this paper.

\begin{figure}[htbp]
    \begin{center}
    \includegraphics[width=1\linewidth]{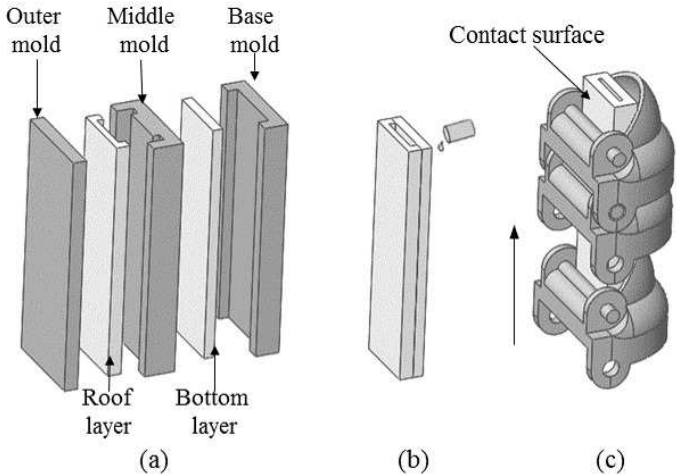}
    \end{center}
    \caption{Fabrication process: (a) Molding of the roof and bottom layer of the soft actuator with a three-part mold, (b) sealing two layers together and (c) assembling the soft actuator with rigid shell segments. }
    \label{fig:FabricationProcess}
\end{figure}

Fig. \ref{fig:FabricationProcess} presents the fabrication process for this hybrid actuator. The casing of the soft chamber was fabricated using a mixture of hyperelastic material including Ecoflex 00-30 and Dragon Skin 30 in a ratio of 6:4. The roof and bottom layers of the soft chamber were made with three molds, including an outer mold, a middle mold and a base mold (Fig. \ref{fig:FabricationProcess}(a)). A thin layer of the same material was used to seal the top and bottom actuators together (Fig. \ref{fig:FabricationProcess}(b)). The soft actuator was dipped into the same mixture to seal one end. A pneumatic tube was fitted to the open end of the soft chamber, and then sealed with silicone proxy. The rigid shell was designed in Fusion 360 and then 3D printed with a Form 2 printer using Clear V2 (FLGPCL02) resin with 0.1mm layer thickness. Three types of shells were designed including the modular shell segment, an end cap to ensure a simple fixture and a distal tip segment for better force measurement at the tip. Finally, the soft chamber was inserted into the 3D printed shells to form the hybrid actuator (Fig. \ref{fig:FabricationProcess}(c)). The contact surface between soft chamber bottom and each shell segment was glued to ensure a consistent stretch along the soft chamber under actuation. 

The hybrid actuator design can be tuned by changing several geometric parameters including internal width and length of the soft actuator, wall thickness, the distance between two axes, shell inner radius and the distance between the joint axis and soft actuator bottom (see Fig. \ref{fig:Bioinspire}). Geometric parameters chosen in this design was based on a further application in a robotic glove. Therefore, several geometrical parameters are determined by the hand size of the researcher involved. 

\section{MODELING OF THE HYBRID ACTUATION AND FORWARD KINEMTACIS}
With the introduction of rigid constraints, computationally inexpensive analytic models and forward kinematics of the hybrid actuator are pursued to estimate and predict its mechanical performance. Once inflated by pressured air, the soft chamber will expand to conform to the internal space of the rigid shells, and the actuator will then either bend or exert force when in contact with external objects. These two mechanisms will be investigated in this part with quasi-static models, and forward kinematics will then be proposed to capture its bending trajectory further. Although there are some simplification and linearization to the geometry transformation, these methods are aimed to demonstrate how soft and rigid components interact, and provide a quick way for performance prediction.

\begin{figure}[htbp]
    \begin{center}
    \includegraphics[width=1\linewidth]{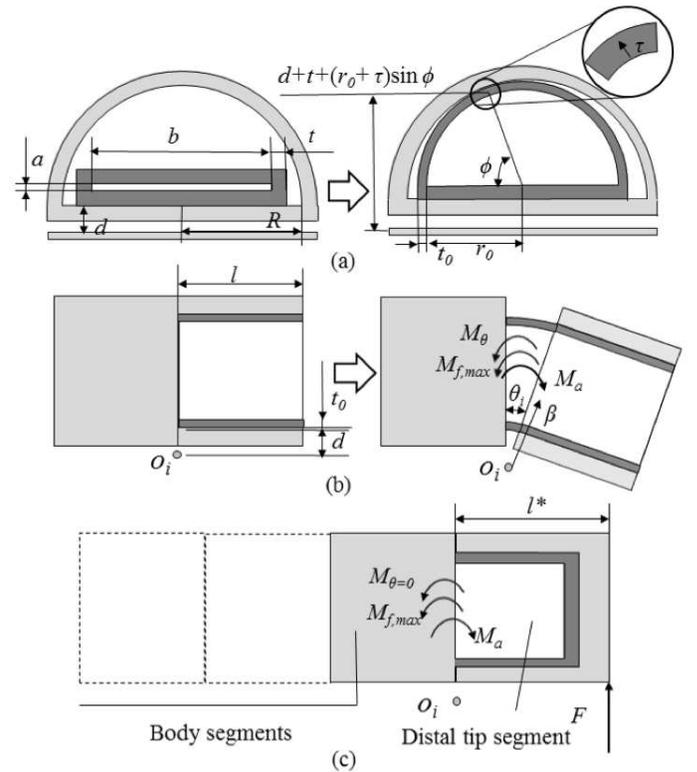}
    \end{center}
    \caption{(a) Expansion process of a hybrid actuator. (b) Soft and rigid interactions and the generated moments. (c) Generated moments when exerting tip force. }
    \label{fig:BendingAnalytic} 
\end{figure}

\subsection{Modeling the Soft Actuator}
In order to fully capture the extension and compression extension phases of the material, Neo-Hookean model was adopted due to its good fit between test data and computed results. Assuming incompressibility, the strain energy is given by
\begin{equation}
    W=\frac{\mu}{2}(I_{1}-1), 
    \setlength\abovedisplayskip{3pt}
    \setlength\belowdisplayskip{3pt}
    \label{eqStrainEnergy}
\end{equation}
where ${\mu}$ is the initial shear modulus and ${I_{1}}$ is the first invariant of three principle stretches (length, width and height) as
\begin{equation}
    I_{1}=\lambda_1^2+\lambda_2^2+\lambda_3^2. 
    \setlength\abovedisplayskip{3pt}
    \setlength\belowdisplayskip{3pt}
    \label{eqFirstInvariant}
\end{equation}
The principle nominal stresses $s_{i} (i=1,2,3)$ can be then obtained as a function of strain energy W, stretches $\lambda_{i}$ and Lagrange multiplier $p$ as
\begin{equation}
    s_{i}=\frac{\partial W}{\partial \lambda_{i}}-\frac{p}{\lambda_{i}}.
    \setlength\abovedisplayskip{3pt}
    \setlength\belowdisplayskip{3pt}
    \label{eqNominalStress}
\end{equation}

When pressured air is pumped in, the soft chamber will first experience an radial expansion process to conform to the internal space of rigid shells with no bending occurring, as shown in Fig. \ref{fig:BendingAnalytic}(a). In width direction, the top and side walls expand into the top wall of a semi-circle with inner radius $r=R-t$, and the average stretch in the top wall can be calculated as $\lambda_{t,2}=(r+t/2)\pi/(b+2t+2a)$. The stretch in length direction of the top wall is negligible due to the constraints from rigid shells before actuating, therefore $\lambda_{t,1}=1$. Stress in length direction will result in an opposing bending torque $M_{\theta}$ around each joint, and the moment equilibrium must be satisfied as:
\begin{equation}
    \left\{
    \begin{aligned}
        & M_{a}=M_{\theta}-M_{geom} (M_{a}<M_{\theta})\\
        & M_{a}=M_{\theta}+M_{f} (M_{a}>M_{\theta}),
    \end{aligned}\right.
    \setlength\abovedisplayskip{3pt}
    \setlength\belowdisplayskip{3pt}
\end{equation}
where $M_{geom}$ is an additional torque generated by geometrical constraint as the hybrid actuator can only bend unilaterally, $M_{f}$ is the torque created by friction, and $M_{a}$ is torque induced by air pressure and can be calculated as

\begin{equation}
    \begin{aligned}
        M_{a} & =\int_{0}^{\pi/2}P_{in}(2r\cos \theta_{i})(r_{0}\sin \theta_{i}+t+d)r_{0}\cos \theta_{i}~d\theta\\
        & =\frac{P_{in}r_0^2(3d\pi+4r_{0}+3\pi t)}{6}
    \end{aligned}
    \setlength\abovedisplayskip{3pt}
    \setlength\belowdisplayskip{3pt}
    \label{TorqueAir}
\end{equation}
where $r_{0}$ is the inner radius of soft chamber when inflated and $P_{in}$ is the relative input pressure. The top wall thickness of the semi-circular soft chamber becomes thinner due to stretch in width direction and $t_{0}=t/\lambda_{t,2}$ is denoted, and therefore $r_{0}=R-t_{0}$.

\subsection{Modeling the Rigid and Soft Interactions}
For simplicity, the geometry of the fractured sphere shell is not considered in this part. After the soft chamber fills up the internal space of rigid segments, the hybrid actuator will start to bend with further increasing pressure. At each bending configuration, there are four bending moments involved in the actuation including the pressure-induced torque $M_{a}$, stretch caused torque of top and bottom wall of the soft chamber $M_{top}$ and $M_{bot}$, and friction-induced torque $M_{f}$ with a maximum $M_{f,max}$. While $M_{a}$ acts counter-clockwise around the joint O,  $M_{top}$, $M_{bot}$ and $M_{f,max}$ all act clockwise around O (see Fig. \ref{fig:BendingAnalytic}).  The moment equilibrium can be described as
\begin{equation}
    M_{a}=M_{\theta}-M_{f}=M_{b}+M_{t}-M_{f}.
    \setlength\abovedisplayskip{3pt}
    \setlength\belowdisplayskip{3pt}
    \label{BendingMomentEqui}
\end{equation}

For the semi-circular soft chamber, stress in length direction of top wall and bottom wall can be calculated in a similar method as in \cite{Polygerinos2015ModelingActuators} as
\begin{equation}
    s_{t,1}=\mu(\lambda_{t,1}-\frac{1}{\lambda_{t,2}^2\lambda_{t,1}^3}),
    \setlength\abovedisplayskip{1pt}
    \setlength\belowdisplayskip{1pt}
    \label{StressTop1}
    \end{equation}
\begin{equation}
    s_{b,1}=\mu(\lambda_{b,1}-\frac{1}{\lambda_{b,1}^3}).
    \setlength\abovedisplayskip{1pt}
    \setlength\belowdisplayskip{1pt}
    \label{StressBot1}
\end{equation}

For the bottom wall, as the stretch in length direction $\lambda_{b,1}$ is a function of the vertical position, a local coordinate $\beta$ is introduced and then it can be written as
\begin{equation}
    \lambda_{b,1}=\frac{\beta \theta_{i}+l}{l}.
    \setlength\abovedisplayskip{1pt}
    \setlength\belowdisplayskip{1pt}
    \label{eq_ASME}
\end{equation}
where $l$ is the distance between two axes and $\theta_{i}$ is the joint angle between two neighboring segments. It follows that
\begin{equation}
    M_{b}=\int_{d+t}^{d} s_{b,1}(b+2t)\beta ~d\beta. 
    \setlength\abovedisplayskip{1pt}
    \setlength\belowdisplayskip{1pt}
    \label{BottomInte}
\end{equation}

For the top wall, two more local coordinates $\phi$ and $\tau$ are introduced to define $\lambda_{t,1}$ as
\begin{equation}
    \lambda_{t,1}=\frac{\left[{l+d+t+(r_{0}+\tau)\sin \phi}\right]\theta_i}{l}
    \setlength\abovedisplayskip{3pt}
    \setlength\belowdisplayskip{3pt}
\end{equation}
so that
\begin{equation}
    M_{t}=2\int_{0}^{t_{0}}\int_{0}^{\frac{\pi}{2}}s_{t,1}\left[(r_{0}+\tau)\sin\phi+t+d\right](r_{0}+\tau)~d\phi d\tau.
    \label{TopInte}
    \setlength\abovedisplayskip{3pt}
    \setlength\belowdisplayskip{3pt}
\end{equation}

A very complicated polynomial solution can be solved for integral \ref{BottomInte}, and integral \ref{TopInte} can only be solved numerically, which would limit the application of this analytic model in a real-time calculation and further analysis. In this study, two approximations are made to obtain an explicit expression of $M_{t}$ and $M_{b}$.

For the hybrid actuator presented, the range of the principle stretch in length direction is approximately 1 to 1.6, in which the nonlinear characteristic of silicone rubber is not obvious. Effective Young’s modulus component $E_{1}$ and $E_{2}$ are introduced to linearize the stress-strain relationship and it is proved in the calibration part to be a reasonable fit. With this approximation, stress component $s_{t,1}$ and $s_{b,1}$ in Eqn. (\ref{StressTop1}) and Eqn. (\ref{StressBot1}) become
\begin{equation}
    s_{t,1}=E_{t,1}(\lambda_{t,1}-1)+E_{t,2},
    \setlength\abovedisplayskip{3pt}
    \setlength\belowdisplayskip{3pt}
    \label{StressTopH}
\end{equation}
\begin{equation}
    s_{b,1}=E_{b,1}(\lambda_{b,1}-1)+E_{b,2}.
    \setlength\abovedisplayskip{3pt}
    \setlength\belowdisplayskip{3pt}
    \label{StressBotH}
\end{equation}
Another approximation was made by replacing the semi-circular top wall with a rectangular wall of the same width and length, and the top of this rectangular is located at a distance $H$ from the joint $O$, which makes $\lambda_{t,1}$ irrelevant with the local coordinates $\phi$ and $\tau$ as
\begin{equation}
    \lambda_{t,1}=\frac{\beta \theta_{i}+l}{l}.
    \setlength\abovedisplayskip{3pt}
    \setlength\belowdisplayskip{3pt}
    \label{StrechTop}
\end{equation}
By substitution of Eqn. (\ref{StressBotH}) into Eqn. (\ref{BottomInte}), and Eqn. (\ref{StrechTop}), (\ref{StressTopH}) into Eqn. (\ref{TopInte}), torques generated by material stretch can be deduced in a simple form which is provided in the appendix part. $H$ can be determined by empirically comparing the value at different joint angles computed with the exact (\ref{TopInte}) and approximation (\ref{TopInteApprox}). For the hybrid actuator in this paper with $(a, b, t, l, R) = (0.5, 10, 1.5, 8, 8) mm$, it is found that approximation expression (\ref{TopInteApprox}) with $H$=7 can accurately capture the relationship between joint angle $\theta_{i}$, ranging from 0 to 30 degrees, and stretch-induced torque of top wall $M_{t}$ with an approximation error less than 4\%. The accuracy of this approximation is further confirmed in calibration part and details are provided in Appendix A. Therefore, using a flax wall with $H=7 mm$ instead of the semi-circular wall of the inflated soft chamber is adopted as a valid approximation of (\ref{TopInte}) and is used to calculate the stretch-induced torque of top wall for all the rest of this study.

\subsection{Modeling Bending and Forward Kinematics}
The hybrid actuator has a hinge-style bending motion, and the total bending angle is the summation of bending angles between two neighboring segments. Due to the relatively small mass of this actuator (less than 18g for an 11-segment configuration), we do not consider the gravity and assume a smooth bending, which is consistent with experimental observations. Combining Eqns. (\ref{TorqueAir}), (\ref{BendingMomentEqui}), (\ref{BottomInteApprox}) and (\ref{TopInteApprox}), the total bending angle can be then expressed as
\begin{equation}
    \begin{aligned}
        \theta &=(n-1)\theta_{i}\\
        &=\frac{K_{1}}{K_{3}}P_{in}-\frac{K_{2}}{K_{3}}
    \setlength\abovedisplayskip{3pt}
    \setlength\belowdisplayskip{3pt}
    \end{aligned}
\label{TotalAngle}
\end{equation}
where ${K_{1}}$, ${K_{2}}$ and ${K_{3}}$ are parameters only related to actuator geometry and friction limit as
\begin{equation}
    \begin{aligned}
        {K_{1}} &=\frac{1}{6}r_0^2(3d\pi+4r_{0}+3\pi t)\\
        {K_{2}} &=\frac{E_{t,2}(r_{0}+t_{0}/2)}{2}\left[H^{2}-{(H-t_{0})}^{2}\right]\\
                & +\frac{E_{t,2}(r_{0}+t_{0}/2)}{2}\left[H^{2}-{(H-t_{0})}^{2}\right]+M_{f,max}\\
        {K_{3}} &=\frac{E_{b,1}(b+2t)}{3l}\left[{(d+t)}^{3}-d^{3}\right]\\ 
                 &+\frac{E_{t,1}(r_{0}+t_{0}/2)}{3l}\left[H^{3}-{(H-t_{0})}^{3}\right].
    \setlength\abovedisplayskip{3pt}
    \setlength\belowdisplayskip{3pt}
    \end{aligned}
    \label{TotalAngle}
\end{equation}
Denote $P_t=[X_t, Y_t, Z_t]^T$ the vector of the actuator tip in the base frame, then the forward kinematics can be given by 
\begin{equation}
    \begin{split}
    \label{FullKinematics}
    P_t=\begin{bmatrix}\sum_{i=1}^{n}l_i\cos (i\theta_{i})\\\sum_{i=1}^{n}l_i\sin (i\theta_{i})\\0 \end{bmatrix},
    \end{split}
\end{equation}
where $l_i$ is the length of link i.

\subsection{Modeling the Blocked-force Test}
An expression to estimate the blocked force can be developed with the previously derived model for bending angle. In this analysis, the hybrid actuator is assumed to be constrained at 0 bending angles. As the pressure is further increased after the soft chamber fills up all the internal space of the rigid segments, the moment equilibrium around the nearest joint to the tip segment must be satisfied as
\begin{equation}
    M_{a}-M_{f,max}-M_{\theta=0}=Fl^{\star}
    \setlength\abovedisplayskip{3pt}
    \setlength\belowdisplayskip{3pt}
    \label{ForceMomentEquilibrium}
\end{equation}
where $M_{\theta=0}$ is the torque induced by the soft material at 0 bending angles, and $l^{\star}$ is the distance between the actuator tip and the nearest joint. 

Combining Eqn. (\ref{TorqueAir}), (\ref{TopInteApprox}),  (\ref{BottomInteApprox}) and (\ref{ForceMomentEquilibrium}), an explicit relationship between output tip force and input pressure can be get as 
\begin{equation}
    \begin{aligned}
        F &=\frac{M_{a}-M_{f,max}-M_{\theta=0}}{l^{*}}\\
          &=\frac{K_{1}}{l_{*}}P_{in}-\frac{K_{2}}{l_{*}}
    \setlength\abovedisplayskip{3pt}
    \setlength\belowdisplayskip{3pt}
    \end{aligned}
    \label{TipForce}
\end{equation}
where ${K_{1}}$ and ${K_{2}}$ are parameters only related to actuator geometry and friction limit, as discussed in section 3.3.

\section{EXPERIMENTAL VALIDATION}
\subsection{Experimental Setup and Calibration}
\begin{figure}[htbp]
    \begin{center}
    \includegraphics[width=1\linewidth]{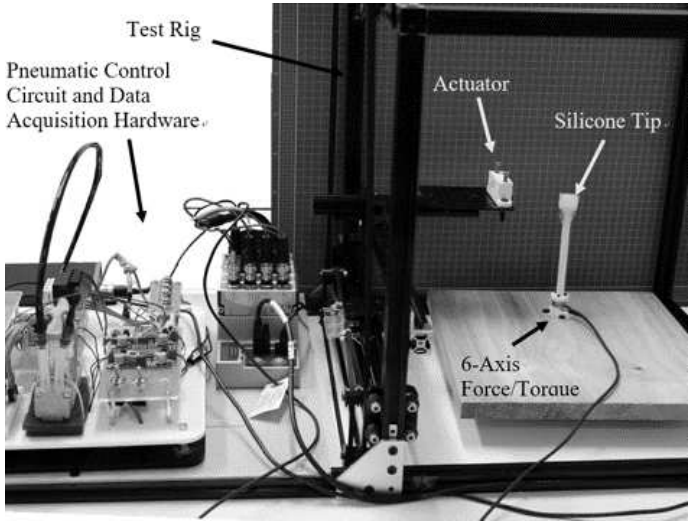}
    \end{center}
    \caption{Evaluation platform consisting of the pneumatic control circuit and data acquisition system, test rig and a 6-axis force/torque sensor.  }
    \label{fig:Platform} 
\end{figure}

To demonstrate the mechanical performance of this proposed hybrid actuator and validate the proposed analytic model, hybrid actuators with six configurations, from 7 to 12 segments, were fabricated and an evaluation platform was developed for the characterization tests. The platform (see Fig. \ref{fig:Platform}) contained a 6-axis force/torque sensor which was mounted onto a three-axis adjustable aluminum frame to measure the force at the tip of the actuator. The measured data was displayed through a graphical interface (GUI) written in LabView. Pressure readings were obtained through a microcontroller (Arduino Mega) and displayed to the user to correspond the pressure in the soft actuator to the force it could produce. The hybrid actuator's proximal tip with air inlet was clamped in a rigid fixture, emulating a boundary condition as fixed. In the bending motion test, the distal tip of the actuator was free to bend in the vertical plane, and a high definition camera was mounted on a tripod to observe the actuators from the side. Post-processing of motion photos was conducted with Adobe Illustrator. A 3D printed post was added to this system mounted on top of the 6-axis force/torque sensor to bring contact with the actuator tip and measure the force exerted. The input pressure was increased gradually, and the force exerted by the tip segment was recorded. 

Calibration was conducted to determine the material parameter $\mu$ and the maximum torque that friction can generate $M_{f, max}$. Three trials of free bending tests with the 8-segment configuration were conducted, and the prediction obtained from the analytic model was compared with experimental data. It was found that the prediction can fit well with the experimental results with $\mu=0.07MPa$ and $M_{f,max}=5N\cdot mm$. The two approximations were also assessed by comparing the exact and approximated values, and the result validated their accuracy as further shown in Appendix A. Although there were two parameters to be calibrated, they were not relevant. As for different configurations, the material used for fabrication and cross-section geometrical characteristics were the same. Therefore, the same $\mu$ and $M_{f,max}$ value were used in all subsequent studies with good results.

\subsection{Comparison between Prediction and Experimental Results}
\begin{figure}[htbp]
    \begin{center}
    \includegraphics[width=1\linewidth]{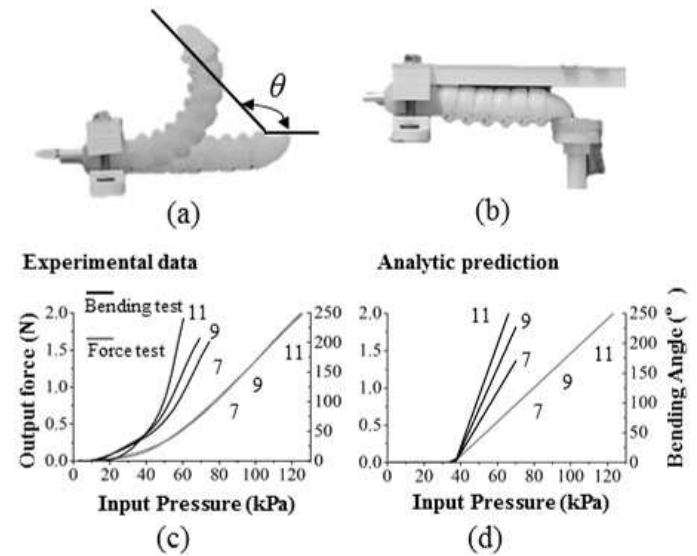}
    \end{center}
    \caption{(a) Bending angle is defined as the deflection angle of the distal tip segment after actuation. (b) Force test set up. (c) Experiment results of bending angle and tip force tests of 7, 9 and 11-segment configuration. (d) Analytic results of bending angle and tip force of 7, 9 and 11-segment configuration.}
    \label{fig:ExperimentalValidation} 
\end{figure}

\begin{figure}[b]
    \begin{center}
    \includegraphics[width=1\linewidth]{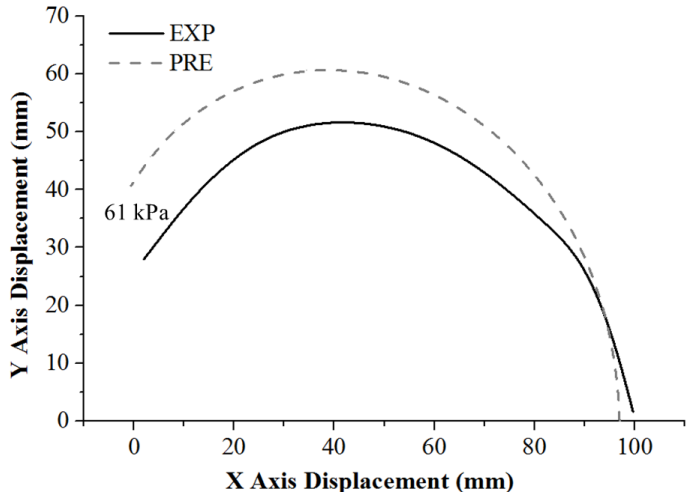}
    \end{center}
    \caption{Comparison of actuator tip trajectories of 10-segment configuration obtained from experimental tests and prediction of forward kinematics.}
    \label{fig:Trajectory} 
\end{figure}

Five different configurations, including 7, 9, 10, 11 and 12 segments, were tested to evaluate the bending capacity. As for each configuration, the proximal tip segment was firmly clamped, hybrid actuators characterized had five different degrees of freedom including 6, 8, 9, 10 and 11. Bending angle is defined as the deflection angle of the distal tip segment as shown in Fig.\ref{fig:ExperimentalValidation}(a). Each actuator was pressured three times to bend in free space to check accuracy, and bending motions were captured by the camera mentioned earlier. Pressure increasing was kept at a sufficiently low rate to ensure the actuator in a quasi-static state. Experimental data was collected to compare with the prediction of the analytic model as shown in Fig. \ref {fig:ExperimentalValidation}(c) and (d), and for a clear reading, only results of  7, 9 and 11 were plotted in this part. Although there are some approximations and linearizations in the analytic model, the findings demonstrate that the analytic model can capture the overall trend of the hybrid actuators. Discrepancies between analytic and experimental results are more evident when total bending angle was small, and this is probably due to not take the geometry of the fractured sphere shells into consideration in the analytic models. Both experimental and analytic results show an enhanced bending capacity with increasing segments, and the 11-segment configuration could reach $250^{\circ}$ total bending angle under about 60 kPa. Fig. \ref{fig:Trajectory} shows the actuator tip trajectories of 10-segment configuration obtained from experimental tests and prediction of the forward kinematics as presented in Eqn. (\ref{FullKinematics}). The position where the proximal tip segment was clamped is denoted the origin (0, 0). It can be seen that the forward kinematics can well describe the trend of the tip trajectory, while some small discrepancies might be caused by the simplifications in analytic models and gravity effects.  

When the free bending is constrained, hybrid actuators are capable of exerting forces either at the interaction points along their body or the tip. In the blocked force test, the proximal tip segments of hybrid actuators were clamped while the distal tips were in contact with the silicone pad connected to the force sensor. A constraining platform was positioned on top of hybrid actuators to ensure that actuators were in 0 bending angles during pressurization (see Fig. \ref{fig:ExperimentalValidation}(b)). In total six configurations were tested including 7, 8, 9, 10, 11 and 12 segments, and experimental data was collected with conforming results. In Fig. \ref{fig:ExperimentalValidation}(c) and (d), results of configurations of 7, 9 and 11 segments are presented. These results have demonstrated the ability of the analytic model to predict the exerting tip force for different configurations, and a clear trend shows that tip force increases with increasing pressure after a certain threshold pressure. Also, it is shown that for hybrid actuators with the same rigid segment and soft actuator design, segment number has a small influence on force generation. The input pressure was increased until 130 kPa which was limited by the valve capacity, and for all configurations, they can reach a maximum tip force of about 2.1 N. 

\section{Conclusion}
In this paper, a hybrid actuator design is proposed by adopting bio-inspirations from lobsters where a soft actuator is enclosed by a series of articulated rigid shells to generate bending actuation. The soft chamber inside can be designed in a simple geometry, and the rigid shells can be 3D printed, which greatly simplifies the fabrication process and enhances the robustness. A folding mechanism is designed in rigid shells so that full protection can be provided to the soft chamber throughout the bending range. Moreover, the rigidly hinged shells make it possible to adopt rigid-body theories to analyze this hybrid actuator with significantly fewer computation costs. Quasi-static models are developed to demonstrate the bending actuation and force exertion of this hybrid actuator, and forward kinematics are given to capture its bending trajectory. Bending tests show that this hybrid actuator exhibits desirable bending motions, forming an enclosed circular workspace. Blocked force tests demonstrate that it can also produce about 2.1N force at the tip at 140 kPa, indicating a comparable output as actuators made from purely soft materials. The experimental data also validates the capacity of proposed analytic models and forward kinematics to capture the overall trend of this hybrid actuator's mechanical performance. Therefore, this lobster-inspired hybridized design offers three major advantages, including reduced complexity in soft chamber design and fabrication, a layer of protection and constraint to the soft chamber inside, and capacity to be easily analyzed with rigid-body kinematics. 

The presented work is still limited in design, experimentation, and analysis to fully characterize its engineering performance. On the application side, it still requires further research and development into engineering applications with such bio-inspired hybrid actuators. For example, the hybrid actuator can be integrated into a wearable glove for hand rehabilitation \cite{Chen2017ARehabilitation} or a robotic hand for adaptive object grasping \cite{Chen2017AComponents}.

\appendix       
\section*{Appendix A}
By substitution of Eqn. (\ref{StressBotH}) into Eqn. (\ref{BottomInte}), and Eqn. (\ref{StrechTop}), (\ref{StressTopH}) into Eqn. (\ref{TopInte}), torques generated by material stretch in section 3.2 becomes
\begin{equation}
    \begin{aligned}
        M_{b} & =\int_{d}^{d+t}(E_{b,1}\lambda_{b,1}+E_{b,2})(b+2t)\beta ~d\beta\\
         & =\frac{E_{b,1}\theta_{i}(b+2t)}{3l}\left[{(d+t)}^{3}-d^{3}\right]\\
        & +\frac{E_{b,2}(b+2t)}{2}\left[{(d+t)}^{2}-d^{2}\right], 
    \end{aligned}
    \setlength\abovedisplayskip{3pt}
    \setlength\belowdisplayskip{3pt}
    \label{BottomInteApprox}
\end{equation}

\begin{equation}
    \begin{aligned}
        M_{t} & =\int_{H-t}^{H}(E_{t,1}\lambda_{t,1}+E_{t,2})(b+2t)\beta ~d\beta\\
         & =\frac{E_{t,1}\theta_{i}(r_{0}+t_{0}/2)}{3l}\left[H^{3}-{(H-t_{0})}^{3}\right]\\
        &+\frac{E_{t,2}(r_{0}+t_{0}/2)}{2}\left[H^{2}-{(H-t_{0})}^{2}\right]. 
    \end{aligned}
    \setlength\abovedisplayskip{3pt}
    \setlength\belowdisplayskip{3pt}
    \label{TopInteApprox}
\end{equation}

\begin{figure}[htbp]
    \begin{center}
    \includegraphics[width=1\linewidth]{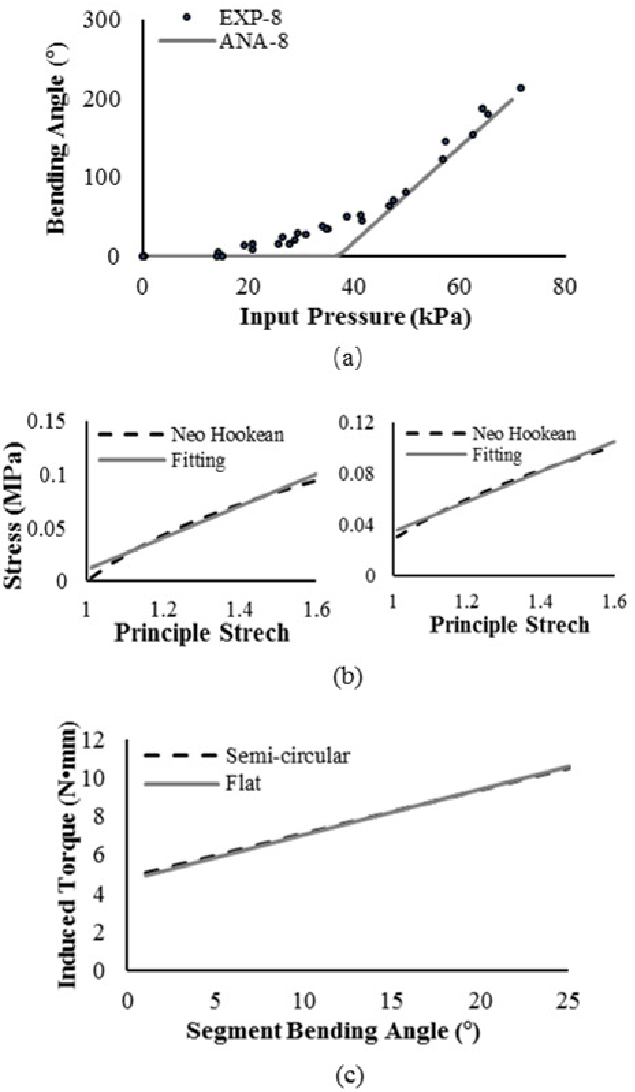}
    \end{center}
    \caption{Calibration process with 8-segment configuration and validation of two approximations. (a) comparisons between experimental results with analytical prediction with $\mu=0.07MPa$ and $M_{f,max}=5N\cdot mm$. (b) Estimation of the accuracy of the approximation of effective Young’s modulus with bottom wall (left) and top wall (right), and the comparison was made with the Neo-hookean model. (c) Estimation of the accuracy of replacing the semi-circular wall with a flat wall with $H=7mm$. }
    \label{fig:CalibrationFigure} 
\end{figure}

For the calibration part in section 4.1, the good match between analytic and experimental results can be seen from Fig. \ref{fig:CalibrationFigure}(a). Accuracy of using Effective Young's modulus components $E_{1}$ and $E_{1}$ to model the non-linear behavior has also been verified in Fig. \ref{fig:CalibrationFigure}(b). Finally, the results of using an flat wall and a semi-circular wall to calculate $M_{t}$ is checked (see Fig. \ref{fig:CalibrationFigure}(c)).  

\bibliographystyle{asmems4}
\bibliography{asme2e}

\end{document}